\documentclass{article}
\usepackage{spconf,amsmath,graphicx}
\usepackage{url}
\usepackage{subfig}
\usepackage{algorithm}               
\usepackage{algorithmic}             
\usepackage{multirow}                
\usepackage{amsmath}
\usepackage{xcolor}
\usepackage{booktabs}
\usepackage{url}
\usepackage{hyperref}
\usepackage{cite}

\title{Towards thinner convolutional neural networks through Gradually Global Pruning}
\name{Zhengtao Wang \qquad Ce Zhu  \qquad Zhiqiang Xia \qquad Qi Guo \qquad Yipeng Liu\thanks{This research is supported by National Natural Science Foundation of China (NSFC, No. 61571102, No. 61602091), the Fundamental Research Funds for the Central Universities (No. ZYGX2014Z003) and National High Technology Research and Development Program of China (863, No.2015AA015903).}}
\address{School of Electronic Engineering / Center of Robotics,\\University of Electronic Science and Technology of China}
\begin{document}
%\ninept
%
\maketitle
\begin{abstract}
Deep network pruning is an effective method to reduce the storage and computation cost of deep neural networks when applying them to resource-limited devices. Among many pruning granularities, neuron level pruning will remove redundant neurons and filters in the model and result in thinner networks. In this paper, we propose a gradually global pruning scheme for neuron level pruning. In each pruning step, a small percent of neurons were selected and dropped across all layers in the model. We also propose a simple method to eliminate the biases in evaluating the importance of neurons to make the scheme feasible. Compared with layer-wise pruning scheme, our scheme avoid the difficulty in determining the redundancy in each layer and is more effective for deep networks. Our scheme would automatically find a thinner sub-network in original network under a given performance.
\end{abstract}
\section{Introduction}
\label{sec:intro}

CNNs have achieved great successes in various pattern recognition tasks, especially in large scale image classification \cite{krizhevsky2012imagenet, simonyan2014very, he2015deep}. However, these deep learning models always contain dozens of layers and millions or billions of parameters. For example, AlexNet \cite{krizhevsky2012imagenet} network contains about 60 millions of parameters, while VGG network contains about 144 millions of parameters. The memory and computation cost of such models are so high that it is difficult to apply them to resource-limited devices such as mobile phones. On the other hand, it turns out that the deep learning models are always over-parameterized \cite{denil2013predicting}, which means that the huge amount of connections of deep learning models could be properly pruned and compressed to reduce the storage and the computation cost. This need has driven the development of the research of deep learning model compression, also known as Deep Compression.

Researchers have proposed various of methods on deep compression. We roughly group them into three categories. \textbf{(1) Approximation} \cite{kim2015compression, denil2013predicting, Novikov2015Tensorizing, denton2014exploiting, zhang2015efficient}: weight matrices and tensors in deep model could be approximated using tensor decomposition techniques. The storage cost of deep model is thus saved. \textbf{(2) Quantization} \cite{Gong2014Compressing, Zhu2016Trained,Amodei2015Deep,Rastegari2016XNOR,Courbariaux2016Binarized,Li2016Ternary,Chen2015Compressing}: by searching or constructing a finite set for candidate parameters, one could map parameters from real number to several candidates. These candidates could be properly encoded later for further compression. The extreme case for quantization is the binary networks, in which the parameters only have two possible values. \textbf{(3) Pruning} \cite{Cun1990Optimal,hassibi1993second,han2015learning,aghasi2016net,Polyak2015Channel}: methods in this category aim to reduce redundant connections, neurons or entire layers of the model. Model pruning could be conducted with different granularities, resulting in reduction in model depth, width or number of connections. Approaches directly training sparse networks are not deemed to be pruning methods because they did not really prune any networks. Compared with directly training network from scratch, deep compression could make the most of existing pre-trained models, which were carefully trained by experts and are efficient in extracting features.

Compared with approximation and quantization in deep compression, model pruning would directly change the structure of the model. The pruned model will have sparse connections, fewer layers or neurons according to different pruning granularities. In practical terms, model pruning at neuron level essentially selects a sub-network in original network, keeping the regularity unchanged. As a comparison, pruning at connection level always results in sparse-connected networks, which not only need extra representation efforts but also not fit well for parallel computation \cite{anwar2015structured}. 

In this paper, we will mainly focus on neuron level pruning, whose aim is to reduce the width of layers in the model. To avoid confusion, we will use the term ``neuron'' to refer to a single neuron in fully-connected layers or a filter in convolution layers. A normal neuron level pruning scheme usually contains three steps: (1) Select neurons to be removed. (2) Drop redundant neurons. (3) Fine-tune the model to recover the performance. This process is usually done in layer-wise. The main disadvantage of the layer-wise scheme is that it is time-consuming and for a given performance target, it is hard to determine how many neurons should be dropped in each layer. We try to solve these problems by pruning the network globally. In each pruning step, all neurons in the network will be taken into consideration at the same time.

The main contributions of this paper are (1) we propose a gradually global pruning scheme for neuron level model pruning, which could automatically find a near-optimal structure for given performance. (2) we propose a simple method to evaluate the contribution score in different layers when selecting redundant neurons. In our scheme, the redundant neurons are selected globally to avoid the difficulty of determining the number of redundant neurons in each layer. In each pruning step, only a small percent of neurons were selected to keep the model stay close to the original local optimal point as far as possible, so that we can recover the performance fleetly through just few epochs of fine-tuning. As a result, one could find a near-optimal structure for specific task and obtain a thinner network, which is particularly suitable for resource-limited devices. Note that the size of the pruned model could be further compressed through other deep compression methods.

In Section 2.1, we will introduce some neuron contribution evaluation methods and adjust them to be compatible with global neuron selection. The gradually global pruning scheme will be given in Section 2.2. Our experiment results with different contribution score evaluation methods and contrast experiments on layer-wise pruning are shown in Section 3. Finally, in Section 4 we make a brief summary of the work and provide some insights for future research.

\begin{figure*}[!h]
\centering
\subfloat[$\overline{R}$ score]{
	\label{fig:original_mean}
	\begin{minipage}[t]{120pt}
		\centering
		\includegraphics[width=4cm, keepaspectratio=True]{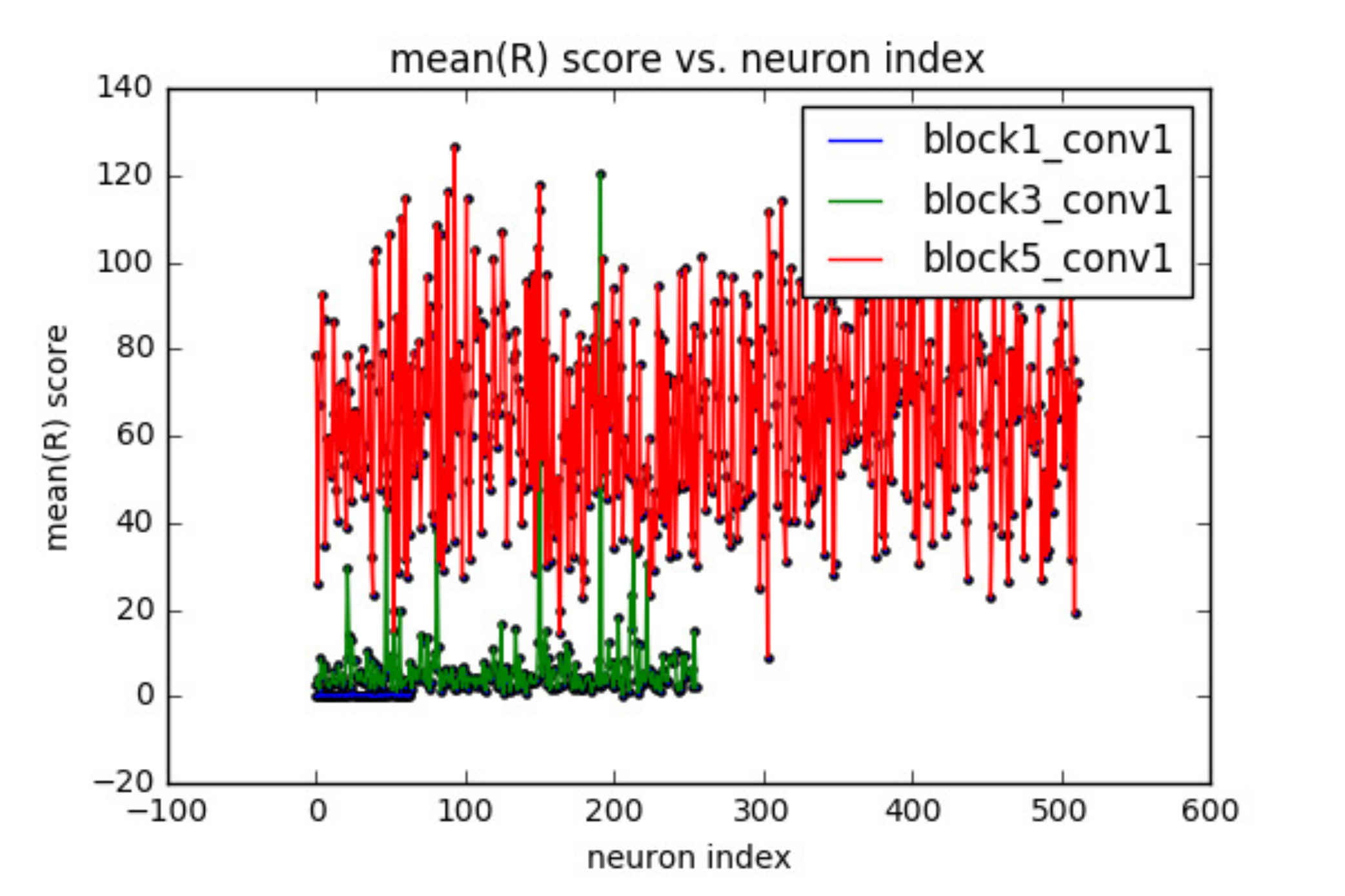}
	\end{minipage}
}
\subfloat[$\sigma(R)$ score]{
	\label{fig:original_sigma}
	\begin{minipage}[t]{120pt}
		\centering
		\includegraphics[width=4cm, keepaspectratio=True]{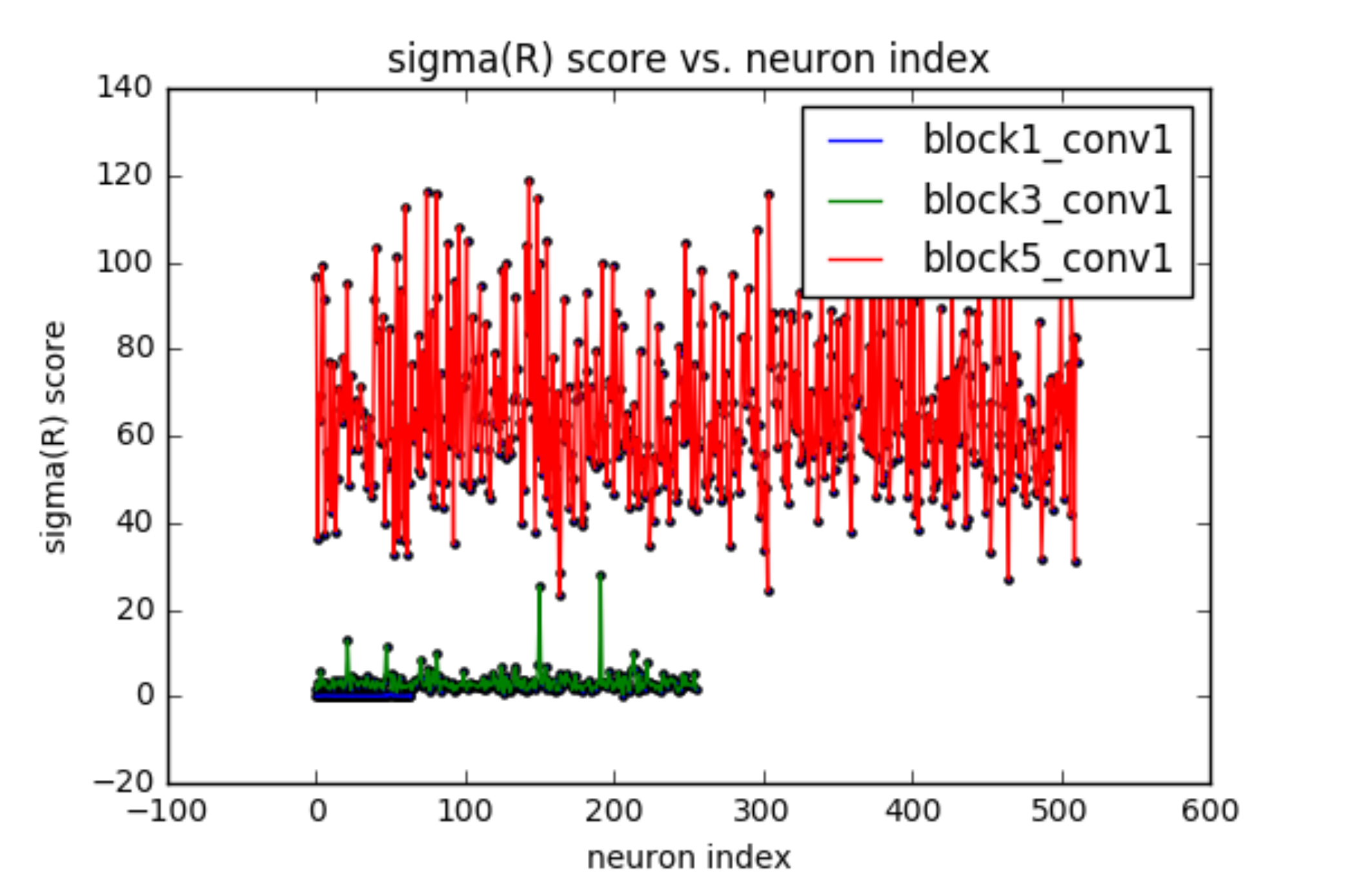}
	\end{minipage}
}
\subfloat[AAWS score]{
	\label{fig:original_AAWS}
	\begin{minipage}[t]{120pt}
		\centering
		\includegraphics[width=4cm, keepaspectratio=True]{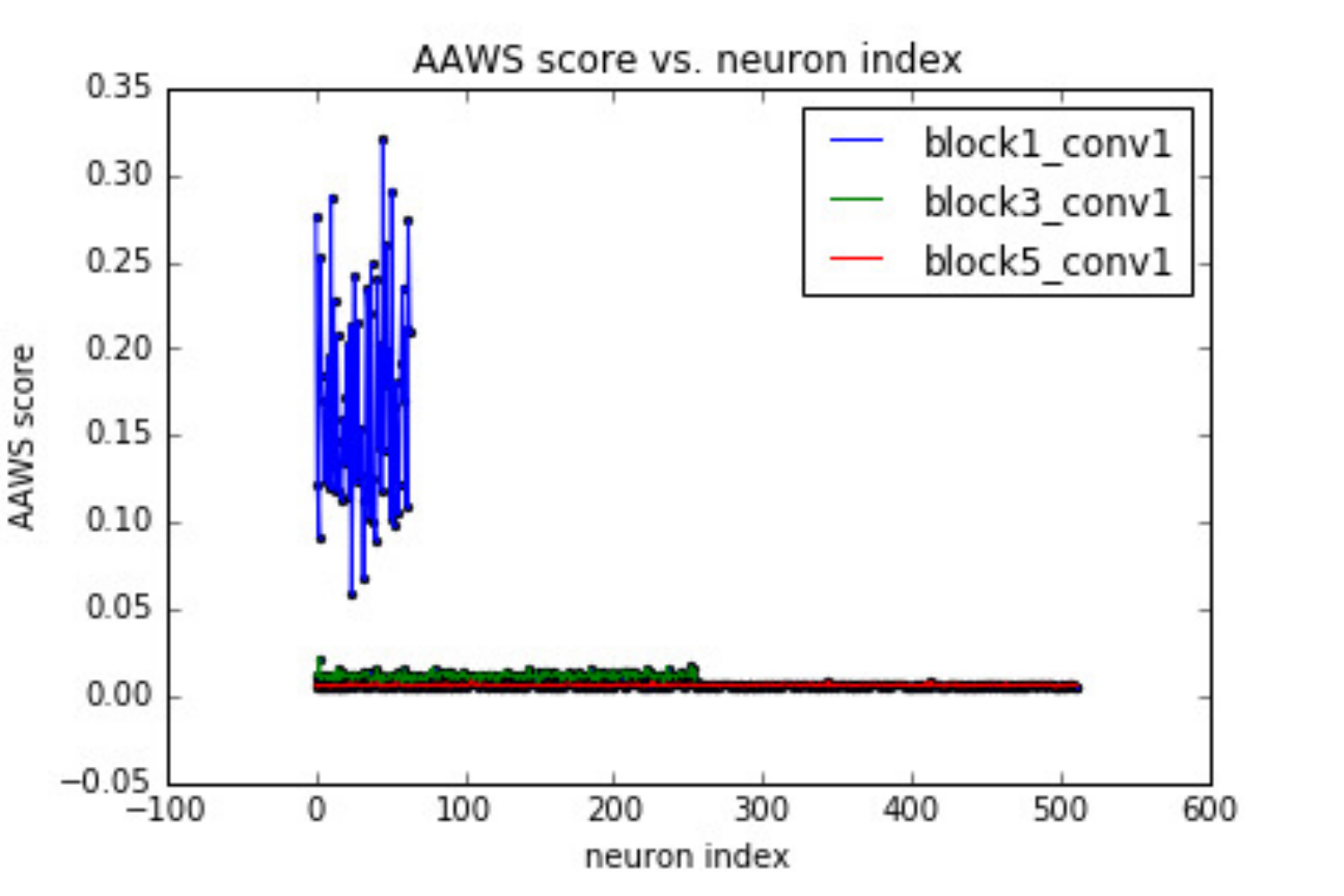}
	\end{minipage}
}

\subfloat[$\overline{R}$ score after modification]{
	\label{fig:adjusted_mean}
	\begin{minipage}[t]{120pt}
		\centering
		\includegraphics[width=4cm, keepaspectratio=True]{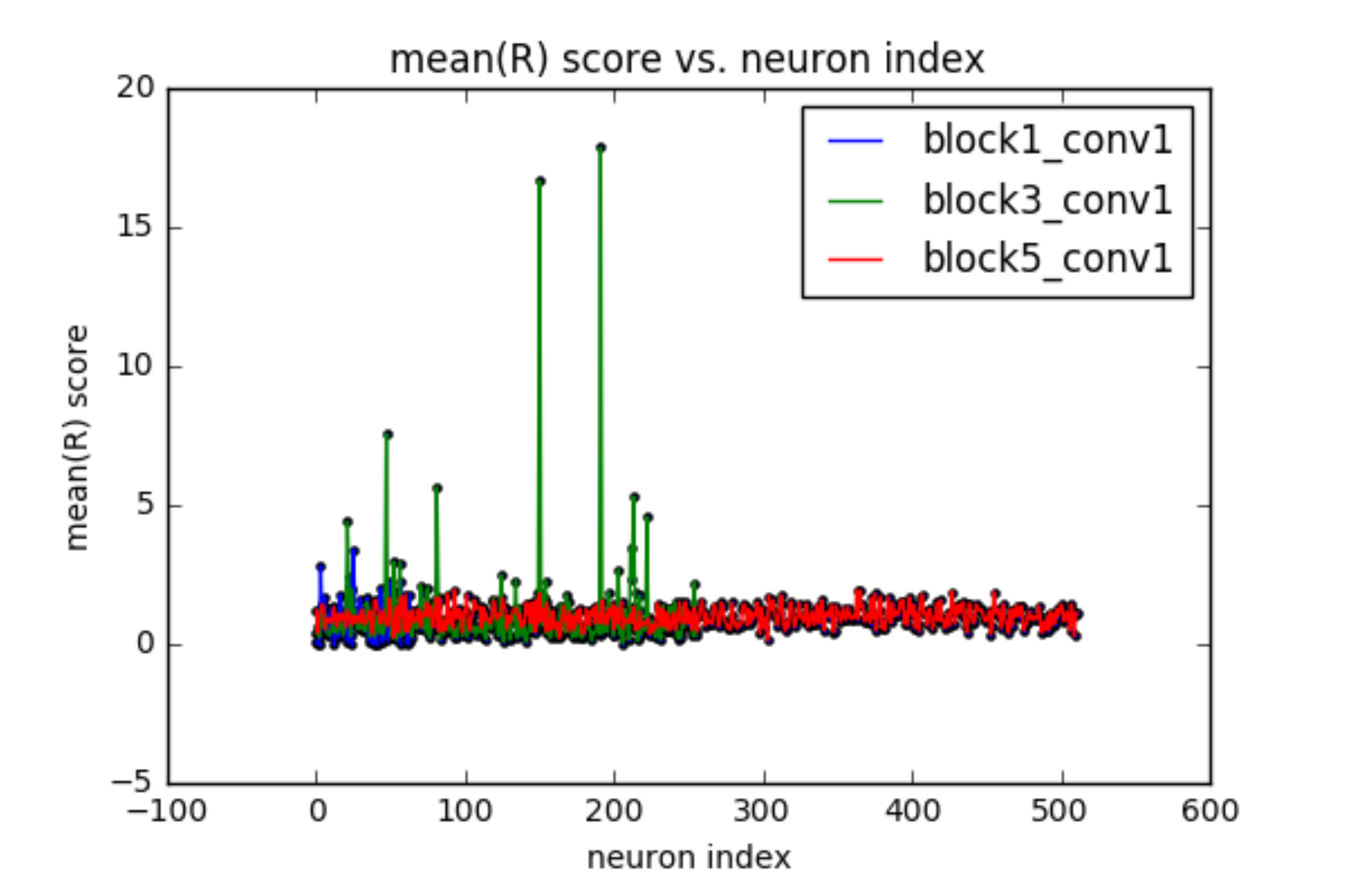}
	\end{minipage}
}
\subfloat[$\sigma(R)$ score after modification]{
	\label{fig:adjusted_sigma}
	\begin{minipage}[t]{120pt}
		\centering
		\includegraphics[width=4cm, keepaspectratio=True]{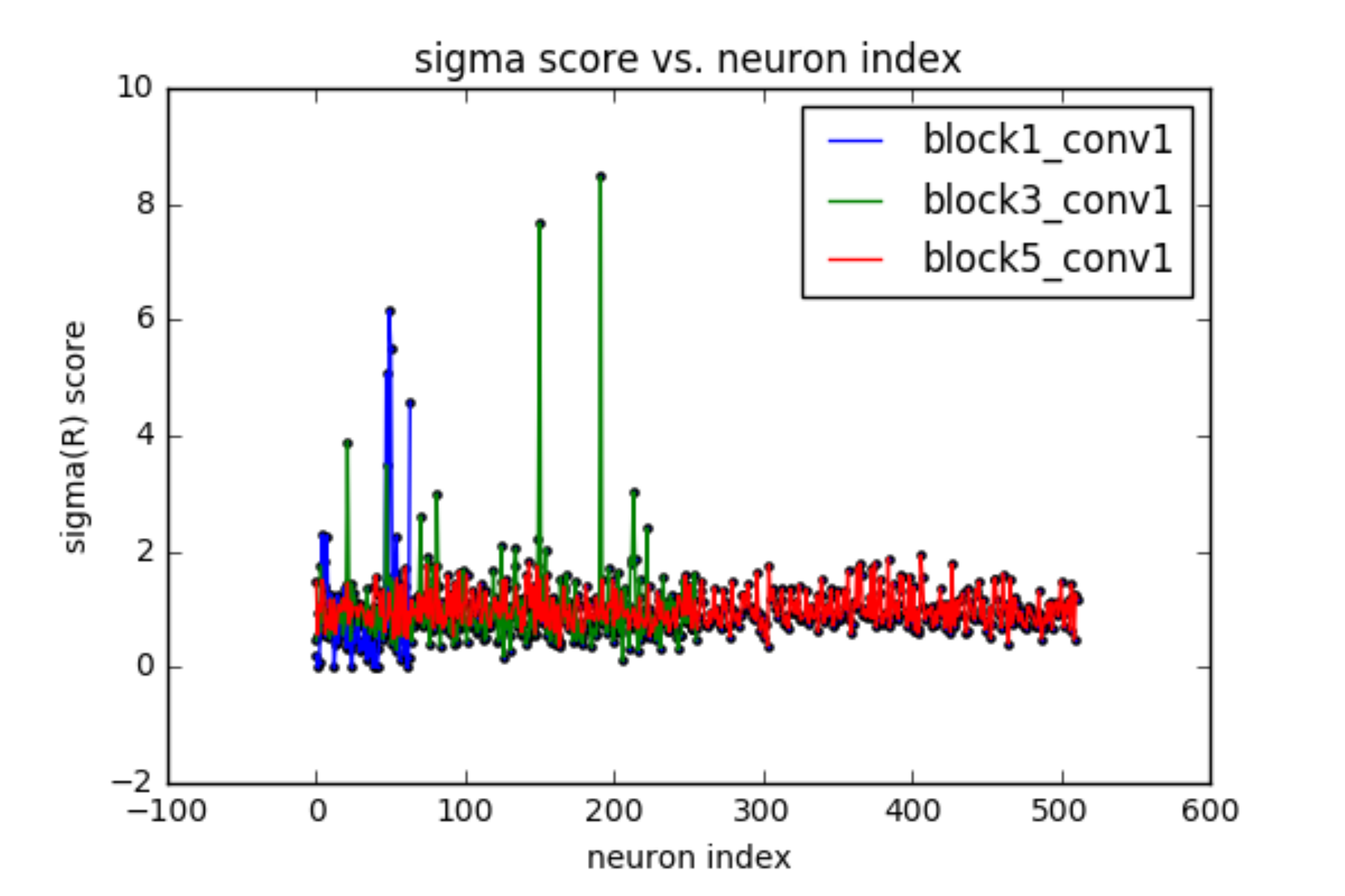}
	\end{minipage}
}
\subfloat[AAWS score after modification]{
	\label{fig:adjusted_AAWS}
	\begin{minipage}[t]{120pt}
		\centering
		\includegraphics[width=4cm, keepaspectratio=True]{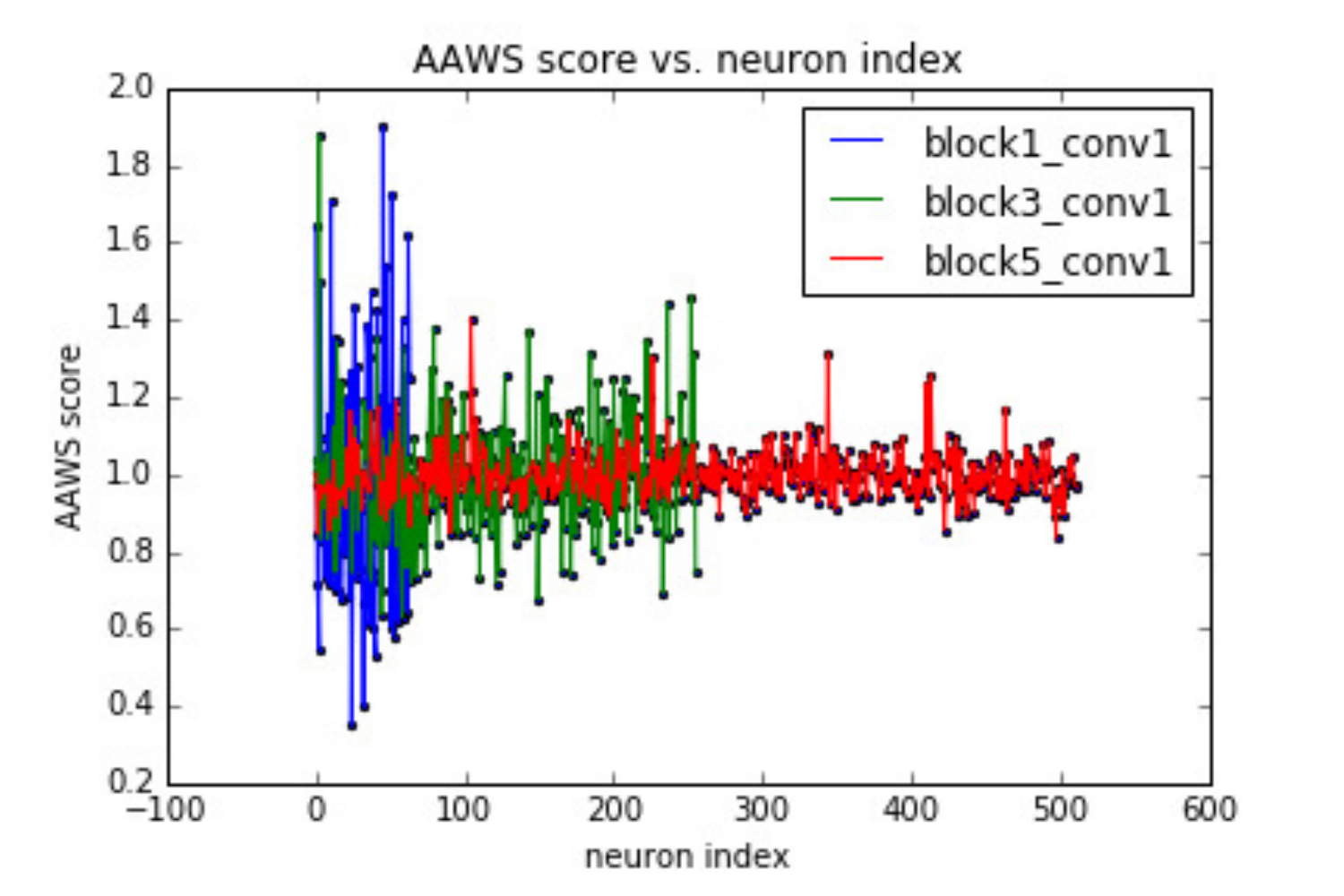}
	\end{minipage}
}

\caption{The score distribution of different metrics}
\label{fig:original}
\end{figure*}

\section{Gradually global pruning scheme}
\label{sec:format}

\subsection{Redundant neurons selection}
\label{ssec:score}
Selecting unimportant or redundant neurons in CNNs is of the prime importance in neuron level pruning. The unimportant neurons have little contribution to the model performance, thus just a few epochs of fine-tuning could compensate the model degradation caused by removing these neurons. Thanks to the huge amount of neurons in deep learning model, neuron selection by try and error is extremely time-consuming. 

 We denote $Y^{l}$ as the output of layer $l$-th for a sample. For convolutional layers, $Y^{l} \in R^{ m\times n\times c}$ is a 3D tensor, where $m$, $n$ and $c$ are feature map width, height and number of input channels, respectively. We define the mean value of $Y^{l}$ as the response of the corresponding filter, then the response of layer $l$ is a vector whose length equals to the number of filters in this layer. For fully-connected layers, the response of each neurons is just the original output value, which is also a vector. We denote the response of a neuron as $R$.

We use 3 contribution score metrics in our scheme. $\sigma(R)$ score is a metric we generalized from \cite{thimm1995evaluating}. The contribution score of neurons is defined as the standard derivation of neuron responses over the training set or a subset sampled from the training set. The idea behind this metric is that a neuron is not important if it has nearly the same output for all training samples. Just like $\sigma(R)$ score, the average responses intensity $\overline{R}$ could also be viewed as a contribution metric if we assume neurons with low average responses intensity are not important. The last metric is generalized from \cite{li2016pruning}. We use the average value of absolute weights sum (AAWS) of a neuron as the contribution score metric. For $i$-th filter at convolutional layer $l$ with $n_p$ parameters, the AAWS score for a filter is defined as:

\begin{equation}Score_{AAWS}(l,i)=\frac{1}{n_p}\sum_{j=1}^{n_p}{|F^{l}_{ij}|}\end{equation}

For fully-connected layers, the AAWS score of a neuron is just the mean of absolute weight sum over all connections starting from that neuron.

These metrics could be directly used in layer-wise pruning scheme because scores within a layer are comparable. However, if we want to conduct a global neuron selection across different layers, they fail to evaluate the neuron contribution properly. Fig.1(a) and Fig.1(b) shows the scores distribution across different layers under $\overline{R}$ and $\sigma(R)$ metrics respectively. We can find that if we take all layers into consideration when selecting redundant neurons globally, neurons at low layers such as conv1\_1 would be treated as the most unimportant neurons and get dropped firstly. However, without low level information extracted by conv1\_1, it is impossible for the network to make inferences. AAWS score in VGG-16 network is shown in Fig.1(c), in this case, the scores in higher layers are more likely to be low. 

According to our observation, The biases show in Fig.\ref{fig:original} are introduced by the position of layers in the model. To make the global neuron selection scheme feasible, we adjust the scores by dividing them by the average score of current layer. After this adjustment, the scores in different layers are mixed up and comparable to each other. As a result, the biases in different layers are eliminated. The modified contribution score can be calculated as:

We make a slight modification on these metrics to make a feasible global neuron selection. We just divided the score of each neuron by the average score of current layer to mix up scores in different layers. By doing this, the biases in different layers are eliminated and the contribution scores of neurons across the model are comparable. 

\begin{equation}Score_{modified}(l,i)=\frac{Score(l,i)}{\frac{1}{N_l}\Sigma_{i=0}^{N_l}Score(l,i)}\end{equation}

$N_l$ is the number of neurons in $l$-th layer. Fig.1(d) to Fig.1(f) show the score distribution after this adjustment.

\subsection{Gradually global pruning scheme}
\label{ssec:framework}

With the neuron selection method proposed in Sec.2.1, we introduce our gradually global pruning scheme in this part. Our scheme is ``global'' because in each pruning step, all neurons in the model instead of just in a layer were taken into consideration. The redundant neurons we selected are the most unimportant neurons in the whole network, not just in a single layer. The global neuron selection method bring us two benefits. Firstly, we do not need to determine how many neurons should be dropped in each layer, which is quite difficult and the redundancy may varies in different layers. Secondly, the number of fine-tunings required is controlled by the given pruning ratio, not the depth of the model, which is particularly useful for pruning deep networks. 

On the other hand, our scheme is ``gradual'' because in each pruning step, only a small percent of neurons were dropped. This is to keep the convergent point after pruning close to the original one as far as possible so that we can recover the performance through just few epochs of fine-tuning. By gradually pruning the network, we can get close to a near-optimal network under a given performance.

In practice, the proposed gradually global pruning scheme prunes a trained network through a ``select-prune-fine-tune'' loop. We summarize the scheme in \textbf{Algorithm 1}. Dropping neurons and updating the network (line 6) could be realized by zeroing the corresponding parameters and updating a mask that forbid the dropped parameters from updating in fine-tuning, or just extract the sub-network that contains important neurons only. In our experiments, we implement model updating in the second way, which could provide model checkpoints additionally.
All experiments were conducted on open source deep learning framework Keras\cite{chollet2015keras} with TensorFlow as the back-end. We will upload our source code to github\footnote{https://github.com/MoyanZitto/GraduallyGlobalPruning} after clear-up for reproduction and future works.

\begin{algorithm}[h]        
\caption{Gradually global pruning scheme.}             
\label{alg:Framwork}                  
\begin{algorithmic}[1]                
\REQUIRE                         
    A trained Model: $M$\newline
    Given performance target: $P_t$\newline
    Contribution score evaluator: $E(\cdot)$\newline
    Pruning ratio generator: $r$\newline
    Training set: $X$\newline
    Validation set: $V$
\ENSURE                           
    A thinner model:  $M$
\STATE Compute the performance $P_m$ of $M$ using $V$
\WHILE {$P_m\geq P_t$}
\STATE Compute the contribution scores of all neurons in $M$ with evaluator $E(\cdot)$
\STATE Sort the scores
\STATE Select $N\times r$ neurons to be prune, where $N$ is the number of neurons in current model
\STATE Drop the selected neurons in the network, get $M_{drop}$, update $M$ by $M_{drop}$ \label{drop}
\STATE Fine-tune $M$ with training set $X$
\STATE Update $P_m$ by the performance of $M$ over $V$
\ENDWHILE
\RETURN $M$               
\end{algorithmic}
\end{algorithm}

\section{Experiments}
\label{sec:pagestyle}

We build a VGG-like network for CIFAR-10 image classification and train it from scratch. The convergent model reaches an accuracy of 87.32\%. The structure of our model is just like VGG-16 model with extra BatchNormalization layers after each convolutional layer and the first fully-connected layer. We set only two fully connected layer before softmax layer. 

In our experiments the pruning ratio is set as 0.05. Pruning ratio is a trade-off parameter that controls the redundancy of pruned model and the speed of the algorithm. A higher pruning ratio will result in more redundant neurons in each step, reducing the number of steps before the program returns. On the contrary, if the pruning ratio is small, the scheme will remove small number of neurons in each step, thus it takes more steps to get close to the optimal structure. In the extreme case, we can remove just one neuron in each step. This pruning ratio can also be determined in some adaptive way.

In the first experiment, we evaluate the performance of different neuron contribution metrics. We force to conduct 7 rounds of pruning without considering the performance degradation. The experiment result are shown in Fig.2(a) and the model structure after 7 rounds of pruning and fine-tuning is shown in Table.\ref{tab:tab1}. According to Fig.2(a), AAWS score has a clear advantage compared with other two metrics. In addition, AAWS score is a date-independent metric, thus we don't need to obtain the statistics of training set in each pruning step. The model structures shown in Table.\ref{tab:tab1} indicate that the AAWS metric tends to give higher score to intermediate layers and lower scores to neurons at lower and higher layers.

\begin{figure}[!h]
\centering
\subfloat[]{
	\label{fig:accuracy}
	\begin{minipage}[t]{120pt}
		\centering
		\includegraphics[width=4.5cm, keepaspectratio=True]{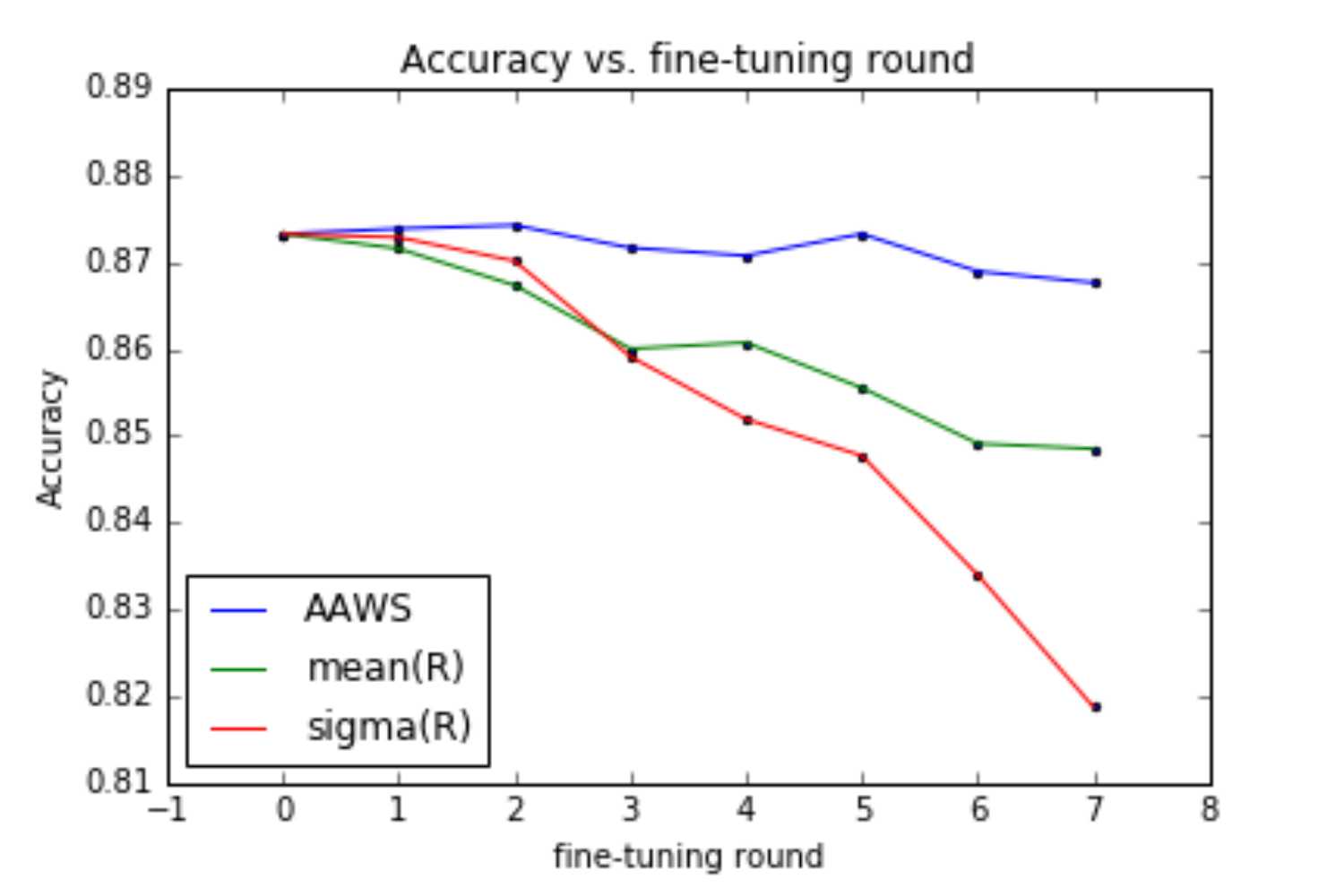}
	\end{minipage}
}
\subfloat[]{
	\label{fig:compare}
	\begin{minipage}[t]{120pt}
		\centering
		\includegraphics[width=4.5cm, keepaspectratio=True]{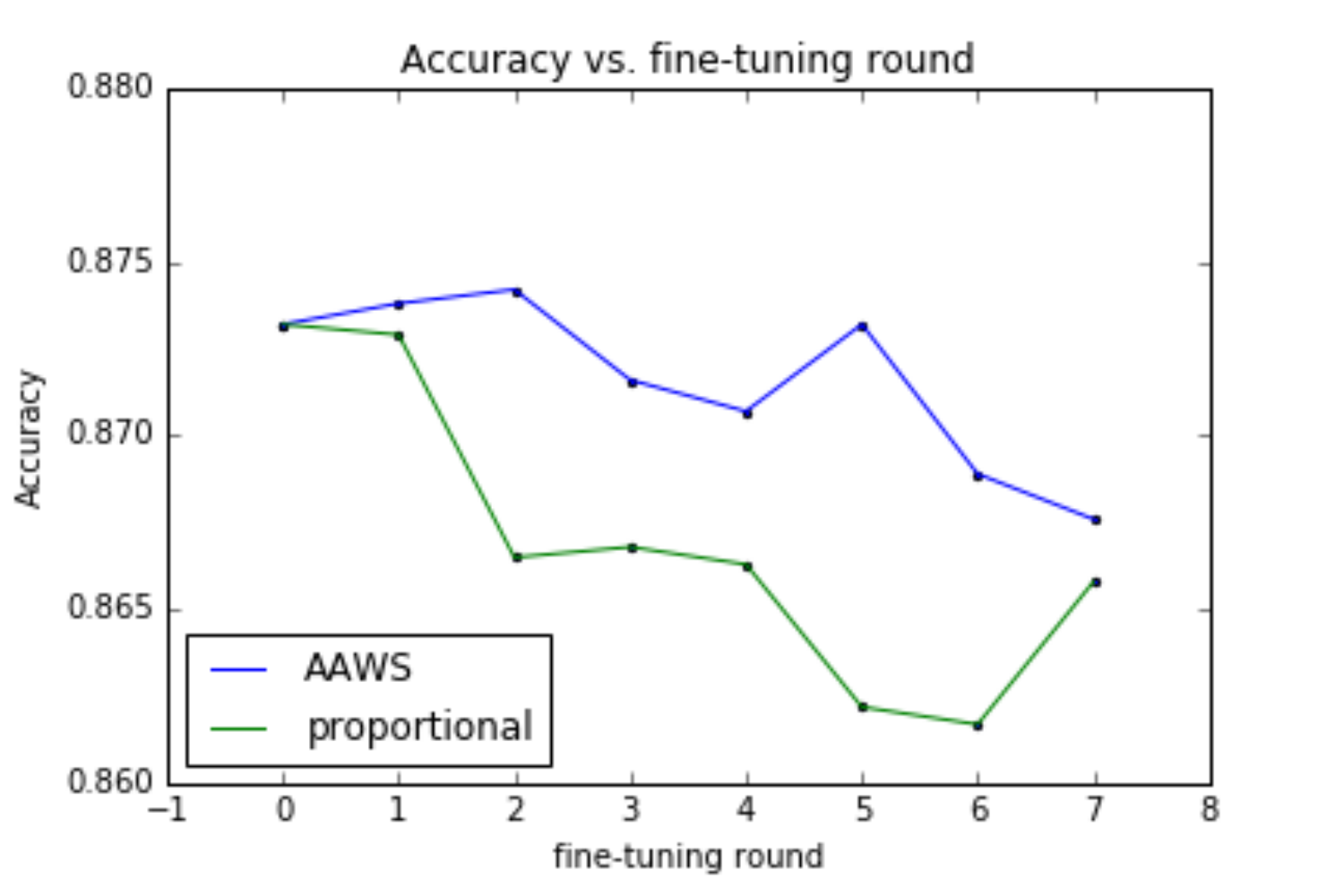}
	\end{minipage}
}

\caption{The performance of different metrics and schemes}
\label{fig:result1}
\end{figure}

We compare our global neurons selection strategy and the layer-wise neurons selection strategy under AAWS score metric. In our second experiment, the neurons were selected at each layer proportionally and all other experiment settings keep unchanged. The results are shown in Fig.2(b). Our global selecting scheme under AAWS score outperforms the layer-wise scheme in every step of pruning. The network structure is shown in the last column of Table.\ref{tab:tab1}.

\begin{table}[!h]
\centering
\caption{}\label{tab:tab1}
\begin{tabular}{cccccc}
\toprule
layer &org.& $\overline{R}$&$\sigma(R)$&AAWS&Prop.\\
\midrule
conv1\_1 & 64 & 35&3&33&45\\
conv1\_2 & 64 & 52&14&34&45\\
conv2\_1 & 128l& 85&70&83&89\\
conv2\_2 & 128 & 72&70&128&89\\
conv3\_1& 256 & 93&168&254&179\\
conv3\_2 & 256 & 173&194&256&179\\
conv3\_3 & 256 & 169&218&256&179\\
conv4\_1 & 512 & 257&314&486&357\\
conv4\_2 & 512 & 405&395&500&357\\
conv4\_3 & 512 & 490&382&448&357\\
conv5\_1 & 512 & 468&452&321&357\\
conv5\_2 & 512 & 436&434&276&357\\
conv5\_3 & 512 & 398&397&229&357\\
fc1 & 512 & 177&199&6&357\\
total & 4736&3310&3310&3310&3304\\
acc.&87.32\%& 84.35\%&81.88\%&86.76\%&86.54\%\\
\bottomrule
\end{tabular}
\end{table}
In the last experiment, we compare the result of our scheme and the layer-wise fine-tuning scheme under the same average pruning ratio. To be precise, in each pruning step we drop about 30.11\% of neurons of a layer and fine-tune the model at once. The accuracy of the model is 86.48\% after 14 rounds of fine-tuning. Note that the number of pruning steps equals to the depth of the model and the optimal pruning ratio for each layer cannot be known in advance. We argue that a similar ``gradually layer-wise pruning'' experiment is no need for conducting because the fine-tuning round of this experiment is extremely large. If the number of layer in the model is $L$ and the pruning round is $n$, the ``gradually layer-wise pruning'' will require $L\times n$ rounds of fine-tunings, while our scheme requires just $n$ rounds.

\section{Conclusion}
\label{sec:typestyle}

In this paper, we propose a gradually global pruning scheme for model compression. By selecting neurons in each pruning step globally, we are able to search a near-optimal structure gradually. As a result, we would obtain thinner networks, which are especially suitable for deep learning application in resource-limited devices. The pruned model inherits the regularity from the original model, thus it is compatible with other deep compression algorithms.
The modification on neuron contribution score is proposed to make the global pruning scheme applicable and is not the optimal. Future works on more suitable contribution score evaluation methods would be very valuable.

% Below is an example of how to insert images. Delete the ``\vspace'' line,
% uncomment the preceding line ``\centerline...'' and replace ``imageX.ps''
% with a suitable PostScript file name.
% -------------------------------------------------------------------------
%\begin{figure}[htb]

%\begin{minipage}[b]{1.0\linewidth}
%  \centering
%  \centerline{\includegraphics[width=8.5cm]{min_mean_distribution.pdf}}
%  \vspace{2.0cm}
%  \centerline{(a) Result 1}\medskip
%\end{minipage}
%

%\begin{minipage}[b]{.48\linewidth}
%  \centering
%  \centerline{\includegraphics[width=4.0cm]{image3}}
%  \vspace{1.5cm}
%  \centerline{(b) Results 3}\medskip
%\end{minipage}
%\hfill
%\begin{minipage}[b]{0.48\linewidth}
%  \centering
%  \centerline{\includegraphics[width=4.0cm]{image4}}
%  \vspace{1.5cm}
%  \centerline{(c) Result 4}\medskip
%\end{minipage}
%
%\caption{Example of placing a figure with experimental results.}
%\label{fig:res}
%
%\end{figure}

%\vfill
%\pagebreak

\bibliographystyle{IEEEbib}
\bibliography{refs}

\end{document}